\documentclass[letterpaper, 10 pt, conference]{ieeeconf}  

\IEEEoverridecommandlockouts                              

\usepackage{graphics} 
\usepackage{epsfig} 
\usepackage{mathptmx} 
\usepackage{times} 
\usepackage{amsmath} 
\usepackage{amssymb}  

\usepackage{algorithm}
\usepackage{algorithmic}

\usepackage{censor}
\usepackage{amsmath,graphicx}
\usepackage[table,x11names]{xcolor}
\usepackage{adjustbox}
\usepackage{multirow}
\usepackage{float} 
\usepackage{arydshln}
\usepackage{makecell}
\usepackage{placeins}
\usepackage{dblfloatfix} 
\usepackage{amssymb}
\usepackage{subcaption}
\usepackage{booktabs}
\usepackage[utf8]{inputenc}
\usepackage[T1]{fontenc}

\usepackage{float}

\title{\LARGE \bf
A Geometric Approach to Soft Actor-Critic with Zonotopes for Locomotion Learning
}

\author{ Panagiotis Roditis$^{1,2}$, Panagiotis P. Filntisis$^{1}$, and Petros Maragos$^{1,2}$%
\thanks{$^{1}$HERON - Hellenic Robotics Center of Excellence, Athena \newline
\hspace*{1.58cm}  Research Center, Greece}%
\thanks{$^{2}$School of Electrical \& Computer Engineering, NTUA, Greece}%
}

\begin{document}

\maketitle
\thispagestyle{empty}
\pagestyle{empty}


\begin{abstract}
Actor--critic methods in reinforcement learning control overestimation bias by taking the
minimum of two critics. This uses the same aggregation rule everywhere,
regardless of how the critics disagree. We propose \textbf{Geometric Zonotope Soft Actor-Critic (GeZo-SAC)}, which
uses auxiliary geometric representations to adapt critic pessimism to the
state and action. Alongside its scalar value, each critic predicts a set of
generators defining a zonotope. Probing this zonotope along sampled directions
provides a geometric width, subtracted from each critic value as a pessimistic offset, and a measure of
disagreement between the two critics, aggregated with log-sum-exp as a smooth approximation to the tropical (max-plus) addition. This
disagreement controls how the critics are combined, moving from a
width-weighted average toward the usual minimum as disagreement increases.
At inference, the deployed policy is an unmodified SAC actor, since the
generators are used only on the critic side during training. Across four MuJoCo-v5 locomotion benchmarks and six off-policy baselines,
GeZo-SAC achieves the highest mean return on Ant-v5 and Hopper-v5 and remains
competitive with other methods on the remaining tasks. Our analysis further shows that
GeZo-SAC achieves the lowest average actuator work and action effort per metre
among the evaluated methods, while maintaining near-zero measured
overestimation frequency across all four environments.
\end{abstract}
\section{Introduction}


Reinforcement learning (RL) has emerged as a powerful framework for
learning control policies for complex robotic systems, enabling agents to
acquire locomotion and manipulation skills directly from interaction with
their environment. In these settings, reliable value estimation is crucial,
as inaccurate predictions of future return can lead to unstable policies,
inefficient motions, or unsafe behavior.

Modern off-policy actor--critic algorithms such as Soft Actor-Critic (SAC) achieve strong
performance in continuous-control tasks due to their sample efficiency, but
their effectiveness depends critically on how the learned $Q$-function guides
policy improvement. In these methods, the critic defines both
Temporal-Difference (TD) targets and the gradients that shape actor updates;
even modest approximation errors can therefore be amplified, as the actor
may preferentially select actions that exploit spurious positive $Q$
errors~\cite{thrun1993issues,pmlr-v119-kuznetsov20a}. This phenomenon
can lead to systematic \emph{overestimation bias}, which can destabilize
training and degrade final policy
quality~\cite{mnih2013playing,fujimoto2018td3,haarnoja2018sac}.


Twin-critic methods mitigate this bias by taking the minimum of two separately
parameterized $Q$ predictions, a strategy adopted by both
Twin Delayed Deep Deterministic Policy Gradient (TD3)~\cite{fujimoto2018td3} and SAC~\cite{haarnoja2018sac}. While effective,
this uses a fixed aggregation rule based only on the two scalar value estimates.
It therefore does not exploit richer structure in how the critics disagree.





We propose a geometry-guided extension of SAC in which each critic predicts,
alongside its scalar value, a set of generators defining a state--action-dependent
zonotope. Directional support probes provide two signals: a geometric band subtracted from each critic value as a pessimistic offset, and a measure of disagreement between the two critics.
This disagreement controls how their pessimistic value estimates are combined,
shifting smoothly from a weighted average toward the minimum as disagreement
increases. See Fig.~\ref{fig:geometric_overview} in
Appendix~\ref{app:geometric_overview} for an illustrated overview and more details of this geometric approach.

Our design draws on connections between tropical (max-plus) geometry and
piecewise-linear function approximators~\cite{maragos2020multivariate,
ChMa17,ChMa18,zhang2018tropical}. Prior work has related Rectified Linear Unit (ReLU) networks
to polyhedral and zonotopic representations, and approximation error to
distances between the associated geometric objects~\cite{misiakos2022hausdorff}.
Motivated by these connections, we use support functions to compare
the learned zonotopes and temperature-controlled Log-Sum-Exp (LSE)
to aggregate directional discrepancies. LSE approaches the maximum,
the addition operator of the tropical (max-plus) semiring, as its
temperature tends to zero.
\newline
Our main contributions are as follows:

\begin{itemize}
    \item We introduce \textbf{Geometric Zonotope Soft Actor-Critic (GeZo-SAC)}, a geometry-guided extension of SAC
    in which each critic predicts an auxiliary zonotope representation.
    Directional support probes provide geometric width and disagreement
    signals that adapt pessimistic value estimates and twin-critic
    aggregation.

    \item We evaluate GeZo-SAC against six off-policy baselines on four
    MuJoCo-v5 locomotion benchmarks. GeZo-SAC achieves the highest mean
    return on Ant-v5 and Hopper-v5 while remaining competitive with other methods
    on the remaining tasks.

    \item We complement return-based evaluation with value-estimation and
    control-quality diagnostics, including overestimation frequency,
    critic--return discrepancy, actuator work, action variation, and
    action effort. GeZo-SAC exhibits near-zero measured overestimation
    frequency and achieves the lowest average actuator work and action
    effort per metre among evaluated methods.
\end{itemize}

\section{Related Work}

\textbf{Overestimation bias and pessimistic value estimation.}
Overestimation bias arises when function approximation and bootstrapping
produce optimistic value errors that the policy can
exploit.  This phenomenon is closely related to the classic
\emph{maximization bias} identified in value-based methods, where using the same
estimator to both select and evaluate actions leads to overestimated targets.
Van Hasselt et al.~\cite{vanhasselt2016double} show that Deep Q-Network suffers from
substantial overestimation on Atari and propose Double Deep Q-Network, which reduces bias
by decoupling action selection (online network) from action evaluation (target
network).  In continuous control, Kuznetsov et al.~\cite{pmlr-v119-kuznetsov20a}
show that even with common heuristics, inaccurate value estimates can be
amplified by policy improvement, degrading stability and performance; they
propose Truncated Quantile Critics (TQC), which models a distribution over
returns and truncates optimistic quantiles (via an ensemble mixture), providing
finer control over overestimation than a simple ``take-the-minimum'' rule. Generalized Pessimism Learning (GPL)~\cite{cetin2023gpl}
introduces an uncertainty-dependent penalty on critic targets and learns
its weight through dual TD-learning, allowing the degree of pessimism
to adapt during training. We compare against its SAC-based instantiation, GPL-SAC. More recently, Twin Distributional Critics with $\lambda$-lower-confidence-bound  (TDC-$\lambda$)~\cite{osman2026tdc}
uses twin distributional critics and selects Bellman targets using a lower confidence bound on the predicted return, with a coefficient $\lambda$
providing explicit control over target conservatism.
Conservative Q-Learning~\cite{kumar2020cql} addresses
overestimation in offline RL by augmenting
the Bellman-error objective with a Q-value regularizer that
encourages conservative estimates. Under stated assumptions,
the expected policy value under the learned critic lower-bounds
the true policy value. This regularization discourages inflated
values for actions outside the dataset distribution, reducing
the incentive to exploit extrapolation errors. Our method
uses support-derived bands and directional disagreement between 
critic zonotopes for adaptive pessimism. The bands
determine value offsets and sample-dependent aggregation weights,
while support discrepancies determine a minibatch-level mixing
coefficient.

\textbf{Uncertainty-aware exploration and critic aggregation.}
Optimistic Actor-Critic~\cite{ciosek2019oac} addresses premature
convergence in SAC by separating conservative policy improvement from
exploration. It uses an approximate lower confidence bound for
policy improvement and an approximate upper confidence bound 
to shift the exploration policy toward actions with high epistemic
uncertainty. Distributional actor--critic methods model the return
distribution and use its statistics, such as variance, to improve
stability and mitigate overestimation. For example, Duan et al.~\cite{duan2025dsact} introduce twin value distribution
learning and variance-aware refinements for more robust critic
updates. SUNRISE~\cite{lee2021sunrise} uses a
Q-function ensemble to estimate uncertainty, reweight Bellman backups,
and select exploratory actions using upper-confidence estimates.
Our approach augments scalar critics with zonotope representations.
Their directional supports provide geometric bands and a measure of
disagreement between these representations. 

\textbf{Tropical geometry and piecewise-linear networks.}
Charisopoulos and Maragos~\cite{ChMa17, ChMa18} and Zhang et al.~\cite{zhang2018tropical} were the first to apply tropical geometry
to the study of deep neural networks, establishing a connection between
feedforward neural networks with ReLU activations and tropical geometry, proving
that such networks are equivalent to families of tropical rational mappings.
Furthermore, they demonstrated that the hidden layers of feedforward ReLU
networks can be represented using zonotopes, which serve as fundamental building
blocks for deeper architectures, and that the decision boundaries of these
networks correspond to tropical hypersurfaces.  Maragos and
Theodosis~\cite{maragos2020multivariate} extended tropical geometrical constructs by
optimally fitting max-plus tropical curves and surfaces to data
approximating polygonal or polyhedral shapes, thereby establishing a link between
tropical geometry and piecewise-linear regression.  
An overview of applications of tropical geometry to machine learning can be found in Maragos et al.~\cite{MCT21}. 
Calafiore et
al.~\cite{calafiore2019lse} introduced a novel class of neural networks called
LSE networks, which serve as universal approximators of convex
functions.  They further showed that difference-LSE networks act as smooth
universal approximators of continuous functions over compact convex
domains~\cite{calafiore2020difference}.  Both LSE and difference-LSE networks
exhibit a close connection to tropical mathematics through the so-called
\emph{dequantization} process, and provide a smooth counterpart to max-type
operators that arise naturally in tropical settings.  Misiakos
et al.~\cite{misiakos2022hausdorff} and Fotopoulos et al.~\cite{fotopoulos2026tropnnc} further showed that, for ReLU networks,
approximation error can be related to Hausdorff-type distances between the
associated polyhedral objects. Our work draws on these connections in practice: we use learned
zonotope generators as geometric representations, directional
support functions to compare these representations, and
temperature-controlled LSE aggregation as a smooth approximation of
the maximum discrepancy over sampled directions. These components
provide practical mechanisms for adaptive pessimism in RL.

\section{Method}

GeZo-SAC augments each SAC critic with a learned geometric
representation and uses disagreement between the two representations
to adapt pessimism. Let $s\in\mathbb{R}^{D}$ denote a state and
$a\in\mathbb{R}^{A}$ denote an action. Critic $i\in\{1,2\}$ outputs a scalar
$Q_i(s,a)$ and $K$ vectors $g_{i,k}\in \mathbb{R}^{D+1}$ collected as 
\[
G_i(s,a) :=
\begin{bmatrix}
g_{i,1}(s,a) & \cdots & g_{i,K}(s,a)
\end{bmatrix}^{\top}
\in \mathbb{R}^{K\times(D+1)}.
\]
The Minkowski sum $\oplus$ of the line segments associated with the
generator vectors $g_{i,k}(s,a)$ defines the \textbf{zonotope}
\[
Z_i(s,a):=\bigoplus_{k=1}^{K}g_{i,k}(s,a).
\]
These line segments are the zonotope's generators, each
contributing to its shape and size (see Fig.~\ref{fig:geometric_overview}a-b).

We use the state feature dimension $D$ to size the geometry in proportion to the observation representation, avoiding a separate dimension hyperparameter. Inspired by representations of extended Newton polytopes  with $D$ slope coefficients and one offset, we adopt $D+1$ auxiliary dimensions for the generators.





Conditioning them on both state and action allows the bands and
aggregation weights to vary across candidate actions at the same
state, providing action-dependent geometric modulation in both
the Bellman target and the actor objective.
\newline
\textbf{Directional support and band.}
For $N=32$ random unit directions
$u_n\in \mathbb{R}^{D+1}$, $n=1 \ldots N$, cached and periodically
resampled during training, we define
\begin{align}
h(u;Z_i(s,a))
&=
\sum_{j=1}^{K}
\left|\langle g_{i,j}(s,a),u\rangle\right|,\\
b_i(s,a)
&=
\frac{1}{N}\sum_{n=1}^{N}h(u_n;Z_i(s,a)).
\end{align}
Here, $h$ is the support function of $Z_i$, defined formally by
$h(u;Z_i):=\sup_{z\in Z_i}\langle z,u\rangle$,
giving its maximum projection onto direction $u$, and $b_i$
estimates its average over the $N$ unit directions.
\newline
\textbf{Generator training signal.}
Generator-head parameters are trained through a squared-band penalty,
a decorrelation penalty on batch-mean generator rows, and a
cross-critic band-symmetry penalty, each added to the critic loss
with weight $10^{-3}$:
\begin{align}
\mathcal{L}_{\mathrm{band}}
&=\mathbb{E}_{\mathcal B}[b_i(s,a)^2],\\
\mathcal{L}_{\mathrm{div}}
&=\frac{1}{K^2}\|C(\bar G_i)-I\|_F^2,
\qquad
\bar G_i=\mathbb{E}_{\mathcal B}[G_i(s,a)],\\
\mathcal{L}_{\mathrm{symm}}
&=\mathbb{E}_{\mathcal B}
[(\tilde b_1-\tilde b_2)^2],
\end{align}
where $C(\cdot)$ is the row-wise cosine-similarity matrix.
The generator heads are trained by geometric regularization,
with TD learning acting indirectly through the shared trunk.

\textbf{Band normalization and pessimistic values.}
Each critic maintains an Exponential Moving Average (EMA) $\bar b_i$ of its batch-mean band.
We define
\[
\tilde b_i(s,a)=\frac{b_i(s,a)}{\bar b_i+\varepsilon},
\qquad
V_i(s,a)=Q_i(s,a)-\kappa_t\tilde b_i(s,a),
\]
treating the EMA normalizers as constants during differentiation.
The effective coefficient $\kappa_t$ is shared between critics and
takes the target- or actor-side form defined below.

\textbf{Band-based actor modulation.}
For actor updates, we apply a second EMA normalization and compute
\begin{align}
\hat b_i
&=\frac{\tilde b_i}{\bar b_i+\varepsilon},&
\hat b_{\mathrm{avg}}
&=\tfrac12(\hat b_1+\hat b_2),\\
\hat b_{\mathrm{diff}}
&=|\hat b_1-\hat b_2|,&
u
&=\frac{\hat b_{\mathrm{avg}}
+\eta\hat b_{\mathrm{diff}}}{\bar s(t)+\varepsilon},\\
c(t)
&=\frac{1}{B}\sum_{m=1}^{B}
\exp[-\tau_c u(s_m,a^\pi_m)],
&&a^\pi_m\sim\pi_\theta(\cdot\mid s_m),
\end{align}
where $\bar s(t)$ is an actor-side EMA of the minibatch
mean of $\hat b_{\mathrm{avg}}$,
$\eta=0.5$, and $\tau_c=1$. The minibatch scalar
$c(t)\in[0,1]$ is detached during the actor update.

\begin{figure*}[t!]
    \centering
    \includegraphics[width=1.0\linewidth]{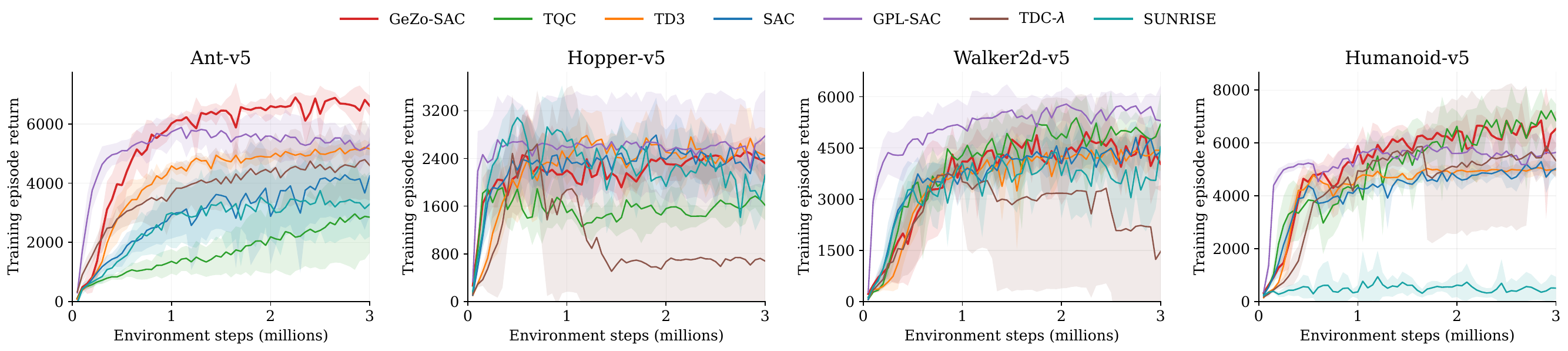}
   \caption{\textbf{Training performance on four MuJoCo-v5 locomotion environments.}
    Learning curves show episodic training returns for Ant-v5, Hopper-v5,
Walker2d-v5, and Humanoid-v5. Per-run returns are smoothed and
interpolated onto a common environment-step grid. Curves represent
the mean over five independent runs ($n=5$), and shaded regions
denote $\pm$ one standard deviation across runs.
GeZo-SAC leads on Ant-v5 and remains competitive on the other tasks.
    }
    \label{fig:training_progress_panel}
    \vspace{-0.4cm}
\end{figure*}

\textbf{Width scheduling.}
Let $t$ denote the number of critic updates.
$\lambda_{\mathrm{base}}(t)$ is zero for the first $10^4$ updates
and increases linearly to $\lambda_{\mathrm{zono}}$ over the next
$2\times10^5$ updates. We define
\begin{align}
\lambda^\pi_t
&=
\mathrm{clip}\!\left(
\lambda_{\mathrm{base}}(t)c(t),0,\lambda_{\mathrm{zono}}
\right),\\
\kappa^\pi_{t,m}
&=\lambda^\pi_t\kappa\beta_{t,m},&
\kappa^{\mathrm{tar}}_t
&=\lambda_{\mathrm{base}}(t)\kappa,\\
\beta_{t,m}
&\sim
\mathrm{Beta}\!\left(1+4c(t),5-4c(t)\right).
\end{align}
During warm-up, the band offsets are zero and the generator penalties
are inactive, although geometric aggregation remains active. Actor and target computations use
$\kappa^\pi_{t,m}$ and $\kappa^{\mathrm{tar}}_t$, respectively.
\newline
\textbf{Batch-adaptive corner directions.}
In addition to $U$, we construct detached directions from the
batch-mean differences between corresponding generator rows:
\begin{align}
\bar{\Delta}g_j
&=
\frac{1}{|\mathcal B|}
\sum_{(s,a)\in\mathcal B}
\bigl(g_{1,j}(s,a)-g_{2,j}(s,a)\bigr),\\
u_j^{\mathrm{corner}}
&=
\frac{\bar{\Delta}g_j}{\|\bar{\Delta}g_j\|_2},
\qquad
\|\bar{\Delta}g_j\|_2>\varepsilon .
\end{align}
This produces at most $K$ additional directions, recomputed
independently for target and actor evaluations. They are concatenated
with the cached probes as
\[
U_\rho=\operatorname{concat}(U,U^{\mathrm{corner}}),
\]
with repeated directions retained.

\begin{algorithm}[t]
\caption{GeZo-SAC training loop (high-level): pessimistic targets with adaptive width.}
\label{alg:gezo}
\begin{algorithmic}[1]
\REQUIRE Actor $\pi_\theta$, twin critics
$\{Q_{\phi_i},G_{\phi_i}\}_{i=1,2}$, targets
$\{\bar Q_{\bar\phi_i},\bar G_{\bar\phi_i}\}_{i=1,2}$,
buffer $\mathcal{B}$
\FOR{each update step}
    \STATE Sample minibatch; sample $a'\sim\pi_\theta(\cdot\mid s')$
    \STATE $V_i\gets\bar Q_i(s',a')-\kappa_t^{\mathrm{tar}}\tilde b_i(s',a')$ \hfill// target values
    \STATE $\rho^{\mathrm{tar}}\gets\exp(-\beta_{\mathrm{geo}}\bar d_{\mathrm{support}}^{\mathrm{tar}})$ \hfill// incl. corner directions
    \STATE $V_{\mathrm{geo}}^{\mathrm{tar}}\gets(1-\rho^{\mathrm{tar}})\min_i V_i+\rho^{\mathrm{tar}}\sum_i w_iV_i$
    \STATE $y\gets r+\gamma(1-d)\left(V_{\mathrm{geo}}^{\mathrm{tar}}-\alpha(0.5)\log\pi_\theta(a'\mid s')\right)$
    \STATE Update critics; adjust $\kappa$ via coverage control
    \IF{policy update step}
        \STATE Compute actor-side $\rho^\pi$ and $V_{\mathrm{geo}}^\pi(s_m,a_{\pi,m})$ using $\kappa_{t,m}^\pi$
        \STATE Update actor and entropy coefficient $\alpha(\cdot)$
    \ENDIF
    \STATE Polyak update targets
\ENDFOR
\end{algorithmic}
\end{algorithm}

\textbf{Geometry-aware support disagreement.}
This construction is applied independently at the two points
in training where critic disagreement is needed: at $(s',a')$
for target computation, and at $(s,a_\pi)$ for the actor update.
This yields direction collections $U_\rho^{\mathrm{tar}}$
and $U_\rho^\pi$ built from the corresponding generators
$\bar G_i(s',a')$ and $G_i(s,a_\pi)$, respectively.
For each sample and direction $u\in U_\rho^{(\cdot)}$,
where $(\cdot)\in\{\mathrm{tar},\pi\}$, define
\begin{align}
\Delta^{(\cdot)}(u)
&=
\big|
h(u;Z_1^{(\cdot)})
-
h(u;Z_2^{(\cdot)})
\big|, \\
s_{\mathrm{scale}}^{(\cdot)}
&=
\tfrac12
\big(
\tilde b_1^{(\cdot)}+\tilde b_2^{(\cdot)}
\big)
+\varepsilon.
\end{align}
We aggregate disagreement using temperature $\tau_d>0$:
\[
d_{\mathrm{support}}^{(\cdot)}
=
\tau_d\log
\sum_{u\in U_\rho^{(\cdot)}}
\exp\left(
\frac{\Delta^{(\cdot)}(u)}
{\tau_d\,s_{\mathrm{scale}}^{(\cdot)}}
\right).
\]
The sum includes every entry in the concatenated direction
collection, including repetitions. These per-sample quantities
are averaged over the minibatch, yielding
$\bar d_{\mathrm{support}}^{\mathrm{tar}}$ and
$\bar d_{\mathrm{support}}^\pi$. We define
\[
\rho^{(\cdot)}
=
\exp\big(
-\beta_{\mathrm{geo}}\,
\bar d_{\mathrm{support}}^{(\cdot)}
\big),
\qquad \beta_{\mathrm{geo}}>0.
\]
Both mixing coefficients are computed without gradient tracking. Figure~\ref{fig:geometric_overview}b illustrates the directional gaps
and their log-sum-exp aggregation into $\rho$.

\textbf{Inverse-band critic weighting and geometric aggregation.}
Given pessimistic value estimates $V_1,V_2$, we define
inverse-band weights
\[
w_i(s,a)
=
\frac{
(\tilde b_i+\varepsilon_w)^{-1}
}{
(\tilde b_1+\varepsilon_w)^{-1}
+
(\tilde b_2+\varepsilon_w)^{-1}
},
\qquad \varepsilon_w=10^{-6}.
\]
We then form
\begin{align}
V_{\mathrm{w}}(s,a)
&= w_1V_1+w_2V_2, \\
V_{\min}(s,a)
&= \min\{V_1,V_2\}, \\
V_{\mathrm{geo}}^{(\cdot)}(s,a)
&= (1-\rho^{(\cdot)})V_{\min}
+\rho^{(\cdot)}V_{\mathrm{w}},
\end{align}
where $(\cdot)\in\{\mathrm{tar},\pi\}$.
The coefficient $\rho^{\mathrm{tar}}$ combines the target critics'
pessimistic value estimates at $(s',a')$, while $\rho^\pi$
combines the online critics' pessimistic value estimates at
$(s,a_\pi)$. In each case, the weights use the corresponding
normalized bands.
\newline
\textit{Intuitive example.} In a two-dimensional example, each generator defines one direction of a centrally symmetric zonotope, and $h(u;Z_i)$ measures its radius along probe direction $u$. Similar supports from the two critics yield small disagreement and favor their weighted average. A large discrepancy in any direction is emphasized by the LSE aggregation, reducing $\rho$ and moving the estimate toward the conservative minimum (see Fig.~\ref{fig:geometric_overview}b-c).
\newline
\textbf{Critic targets.}
Target critics compute $V_{\mathrm{geo}}^{\mathrm{tar}}(s',a')$
using $\kappa_t^{\mathrm{tar}}$. For $a'\sim\pi_\theta(\cdot\mid s')$, the Bellman target is
\[
y=r+\gamma(1-d)
\left[
V_{\mathrm{geo}}^{\mathrm{tar}}(s',a')
-\alpha(0.5)\log\pi_\theta(a'\mid s')
\right].
\]
Here, $d=1$ for true termination and $d=0$ otherwise;
truncation alone does not stop bootstrapping.
The target is computed without gradient tracking.

The base coefficient $\kappa$ is adapted through empirical
target-inclusion control. A center-only target $y_{\mathrm{ctr}}$
uses the same backup with $\min_i\bar Q_{\bar\phi_i}(s',a')$
as its value term. When $\lambda_{\mathrm{base}}(t)>0$, the
controller measures the fraction of these targets contained in
\[
\frac{Q_1(s,a)+Q_2(s,a)}{2}
\ \pm\
\frac{\lambda_{\mathrm{base}}(t)\kappa}{2}
\bigl(\tilde b_1(s,a)+\tilde b_2(s,a)\bigr).
\]
An EMA of this fraction guides the clipped log-space update of
$\kappa$ toward a target inclusion rate of $0.8$.
This target-inclusion rate is used only to adapt $\kappa$ and should
not be interpreted as statistical coverage of the true policy value.
\newline
\textbf{Additional critic regularization.}
Each critic loss includes squared penalties
on value magnitudes exceeding an adaptive reward-based
threshold and on positive TD residuals, weighted by
$10^{-4}$ and $10^{-3}$, respectively.
\newline
\textbf{Actor objective.}
For each minibatch sample $m$, the online critics compute
pessimistic value estimates at $(s_m,a_{\pi,m})$ using
$\kappa^\pi_{t,m}$. Geometric aggregation with $\rho^\pi$
produces $V_{\mathrm{geo}}^\pi(s_m,a_{\pi,m})$.
The minibatch actor loss is
\[
J_\pi
=
\frac{1}{B}\sum_{m=1}^{B}
\left[
\alpha_t^\pi(\beta_{t,m})
\log\pi_\theta(a_{\pi,m}\mid s_m)
-
V_{\mathrm{geo}}^\pi(s_m,a_{\pi,m})
\right],
\]
where $a_{\pi,m}\sim\pi_\theta(\cdot\mid s_m)$.
The actor uses $\alpha_t^\pi(\beta)=e(t)\alpha(\beta)$,
with $e(t)=1.5$ for $t<5\times10^3$ critic updates
and $e(t)=1$ thereafter.
The $\beta_{t,m}$ are sampled conditionally on the modulation
scalar $c(t)$ as defined above. Higher $c(t)$ increases their
conditional mean.
\newline
\textbf{Update schedule.}
At each learner update, we update the critics and apply
Polyak averaging to the target networks. The coverage
controller is updated when active.
The actor and entropy-coefficient network are updated every
second critic update; $t$ indexes critic updates throughout.
The actor-side quantities $c(t)$, $\lambda^\pi_t$,
$\beta_{t,m}$ and $\rho^\pi$ are recomputed only at policy
updates.
\newline
\textbf{Entropy coefficient.}
A small network $f_\psi$ defines
$\alpha(\beta)=\exp(\mathrm{clip}(f_\psi(\beta),-3,2))$
and is optimized using
\[
\begin{aligned}
J_\alpha
=
-\frac{1}{B}\sum_{m=1}^{B}
\log\alpha(\beta_{t,m})(\hat\ell_m-A),
\qquad
\hat\ell_m
=
\log\pi_\theta(a_{\pi,m}\mid s_m).
\end{aligned}
\]
where the target entropy is $-A$. The sampled
$\beta_{t,m}$ and policy log-probabilities are treated as
constants during this update. Critic targets use $\alpha(0.5)$,
whereas actor updates use $\alpha_t^\pi(\beta_{t,m})$ with the
early entropy multiplier defined above.

Algorithm~\ref{alg:gezo} summarizes the training procedure.


\begin{table*}[t]
\centering
\small
\setlength{\tabcolsep}{4pt}
\begin{tabular}{lccccccc}
\toprule
Environment & GeZo-SAC (ours) & TQC & TD3 & SAC & GPL-SAC & TDC-$\lambda$ & SUNRISE \\
\midrule
Ant-v5 & \textbf{7.17(0.21)} & 4.45(2.05) & 5.77(0.23) & 4.43(1.24) & \underline{6.09(0.52)} & 5.39(1.53) & 5.32(0.77) \\
Hopper-v5 & \textbf{3.77(0.11)} & 3.44(0.38) & 3.52(0.09) & 3.51(0.17) & 3.38(0.21) & 3.59(0.15) & \underline{3.63(0.08)} \\
Humanoid-v5 & \underline{7.89(0.58)} & \textbf{8.61(0.25)} & 5.23(0.20) & 5.60(0.09) & 6.13(0.66) & 6.85(0.54) & 0.62(0.08) \\
Walker2d-v5 & 5.74(0.30) & \textbf{6.22(0.41)} & 4.61(0.52) & 5.25(0.47) & \underline{6.06(0.47)} & 4.99(0.85) & 5.79(0.44) \\
\bottomrule
\end{tabular}
\caption{Mean evaluation return (in thousands) across five training runs, using the best checkpoint from each run and 300 deterministic evaluation episodes. Parentheses show the sample standard deviation. Bold and underline denote the best and second-best means. GeZo-SAC uses $K=15$ in all environments.}
\label{tab:best300-7-methods}
\end{table*}

\begin{table}[t]
\centering
\footnotesize
\setlength{\tabcolsep}{3pt}
\begin{tabular}{lccccc}
\toprule
Method & \makecell{Work/m $\downarrow$} & \makecell{Dist. $\uparrow$} &
\makecell{TV/m$ \downarrow$} & \makecell{Effort/m $\downarrow$} &
\makecell{Return $\uparrow$}\\
\midrule
SAC             & 1.028 & 0.817 & 1.266 & 1.100 & 4.70 (0.32)\\
TD3             & 1.275 & 0.786 & 1.951 & 1.936 & 4.77 (0.16)\\
TQC             & 1.137 & \textbf{1.283} & 0.941 & 0.937 & \underline{5.68 (0.51)}\\
GPL-SAC         & \underline{0.934} & 0.988 & \underline{0.901} & \underline{0.926} & 5.42 (0.23)\\
TDC-$\lambda$             & 1.068 & 1.002 & 1.463 & 1.441 & 5.21 (0.44)\\
SUNRISE         & 1.176 & 0.751 & 1.890 & 2.551 & 3.84 (0.23)\\
GeZo-SAC (ours) & \textbf{0.878} & \underline{1.242} & \textbf{0.837} &
                  \textbf{0.763} & \textbf{6.14 (0.17)}\\
\bottomrule
\end{tabular}
\caption{\textbf{Mechanical metrics and return across the four locomotion tasks.}
Values are averaged across environments; mechanical metrics are
normalized within each environment by its median. Lower is better for Work/m, TV/m, and Effort/m;
higher is better for Distance and return. Bold and underline denote the
best and second-best values.}
\label{tab:mechanical_averages}
\end{table}

\begin{figure}[t]
\centering
\includegraphics[width=\linewidth]{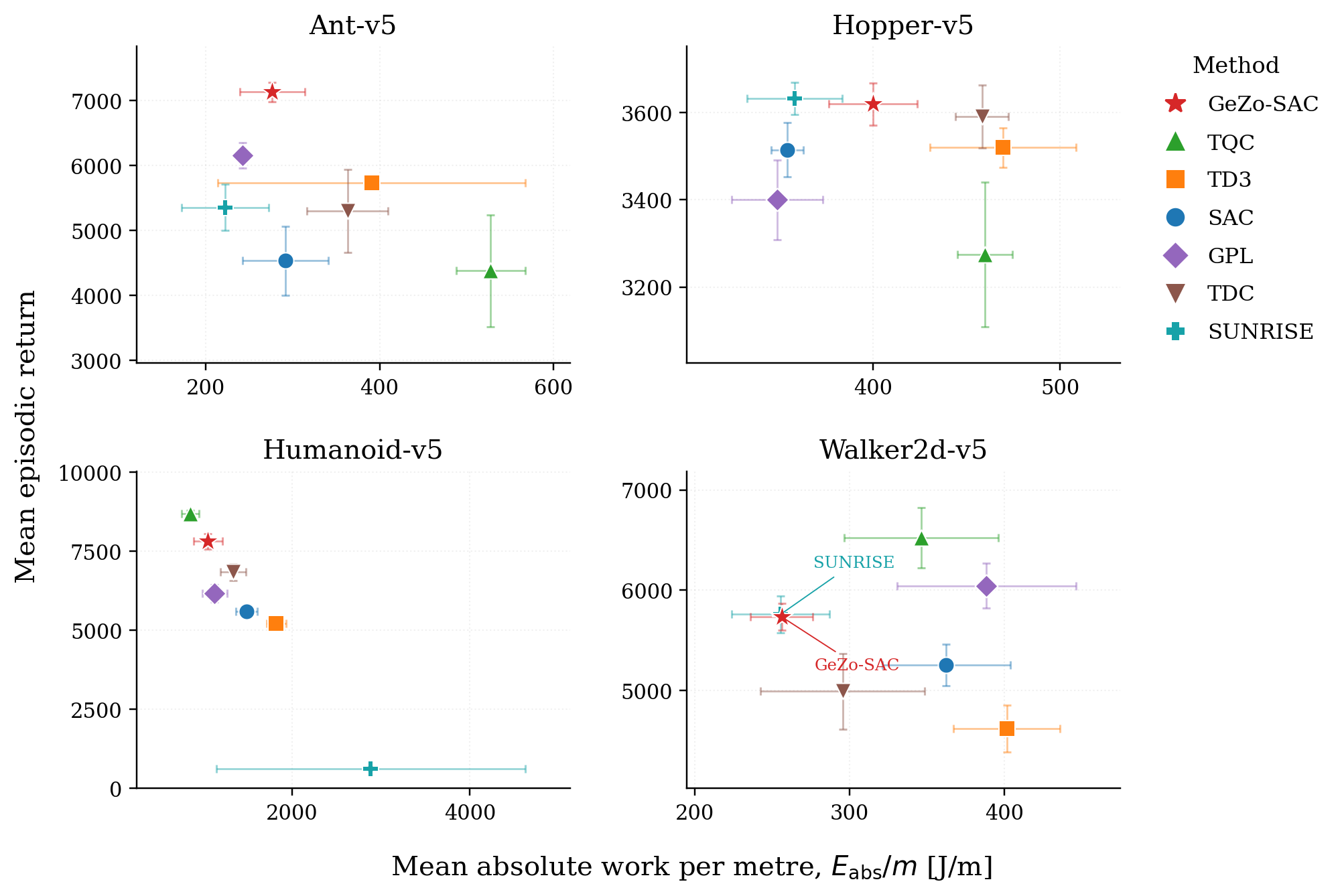}
\caption{\textbf{Performance--energy trade-off.} Mean episodic return versus mean absolute mechanical work per metre, $E_{\mathrm{abs}}/m$ (J/m), across four locomotion tasks. Each marker represents one method; points toward the upper left combine higher return with lower work per metre.}
\label{fig:oneplot_all_envs}
\end{figure}

\begin{figure}[t]
\centering
\includegraphics[width=.80\linewidth]{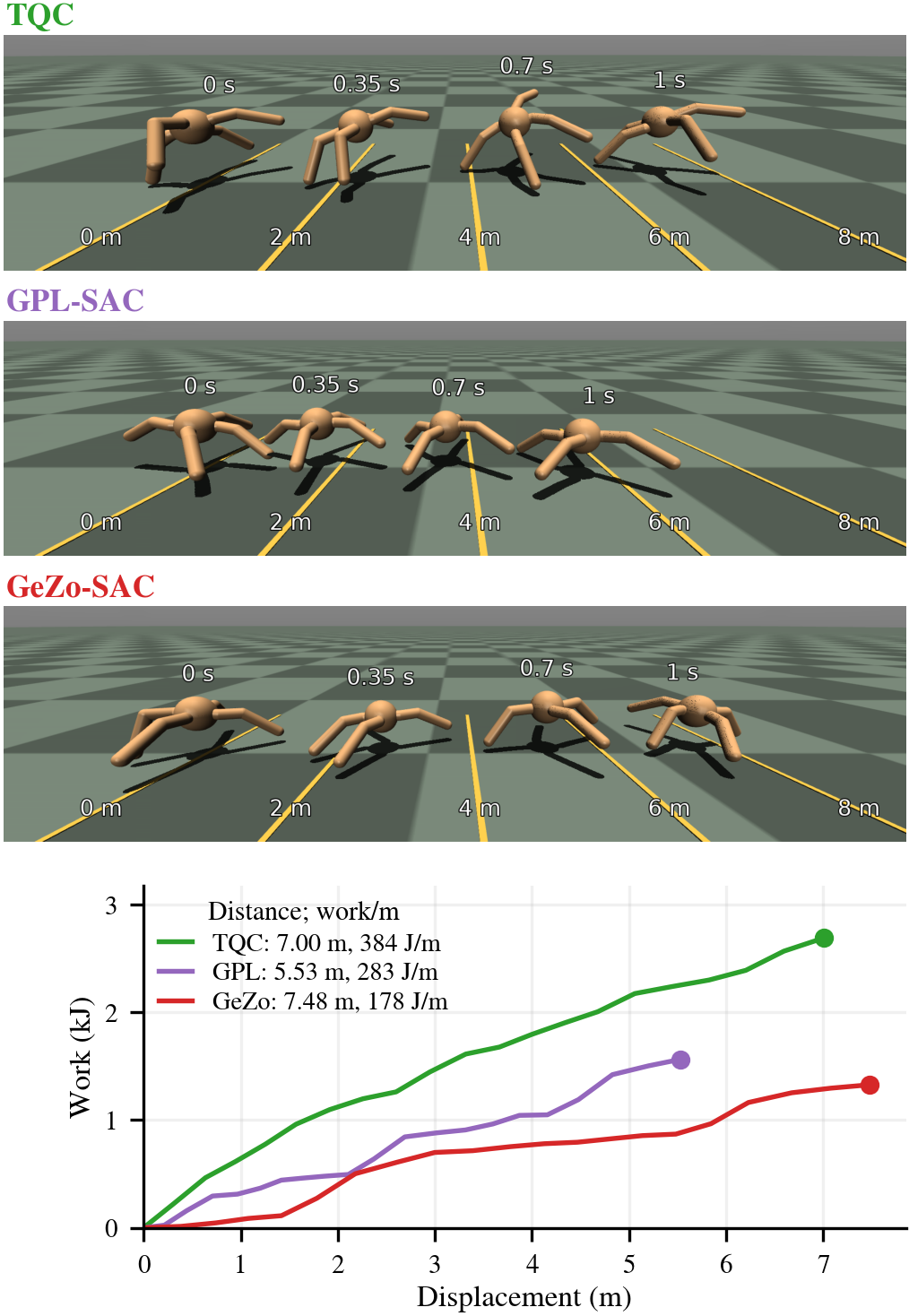}
\caption{\textbf{Locomotion and mechanical cost on Ant-v5.}
\textbf{Top:} Overlaid poses of TQC, GPL-SAC, and GeZo-SAC at
matched times over a one-second interval, shown at a common spatial
scale.
\textbf{Bottom:} Cumulative actuator work versus forward displacement over the
same interval. GeZo-SAC travels farther while requiring less work per
metre in these illustrative rollouts.}\label{fig:ant_rollout}
\vspace{-0.35cm}
\end{figure}

\begin{figure}[t]
    \centering
\includegraphics[
    width=\linewidth,
    trim=0 30pt 0 0,
    clip
]{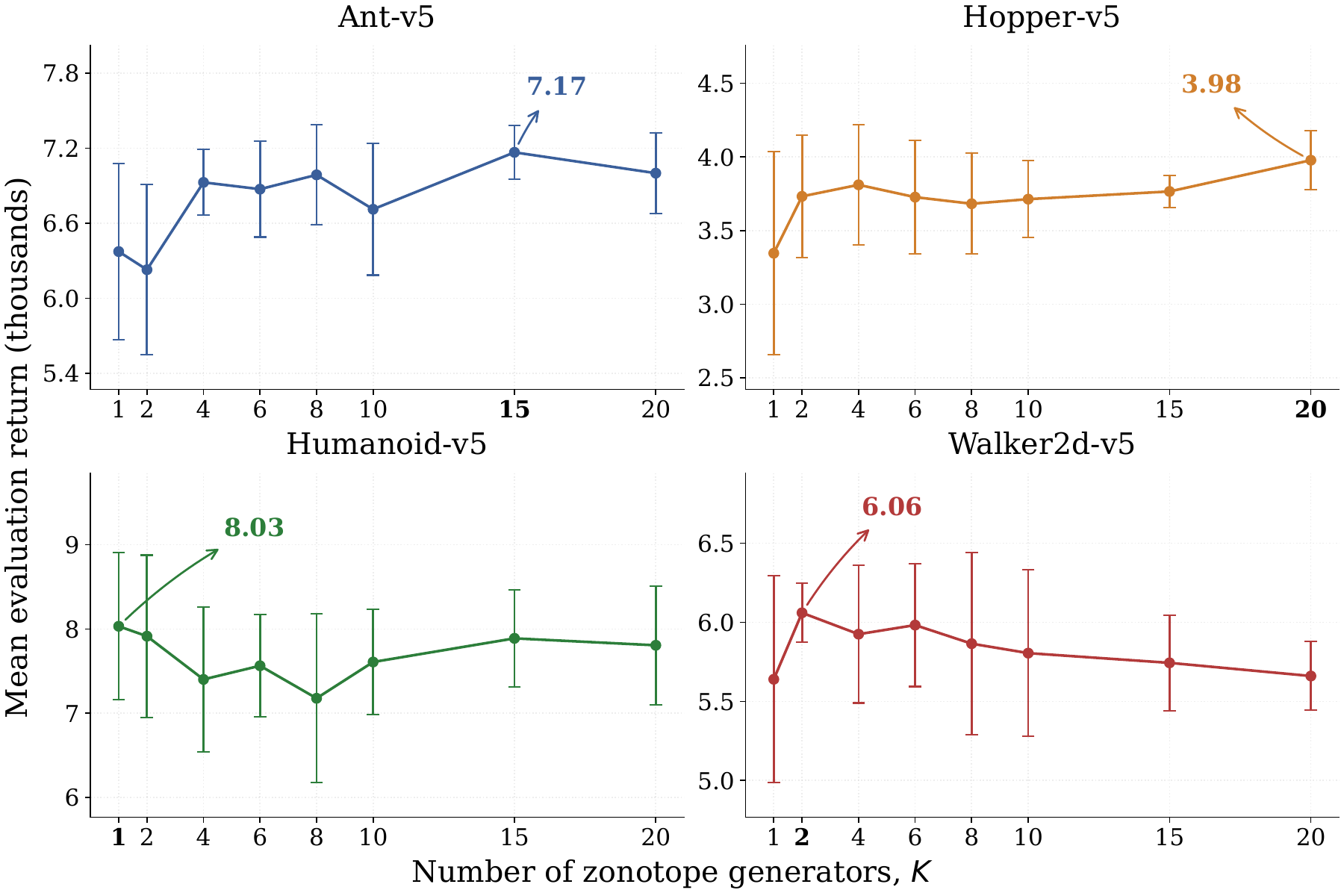}
\caption{\textbf{Sensitivity to the number of zonotope generators.}
    Curves show mean evaluation return over five training seeds,
    with shaded regions and error bars indicating $\pm$ one standard
    deviation. Bold labels mark the highest observed mean in each
    environment.}
    \label{fig:zono_comparison}
\end{figure}

\begin{figure*}[t!]
    \centering   
    \includegraphics[width=\linewidth]{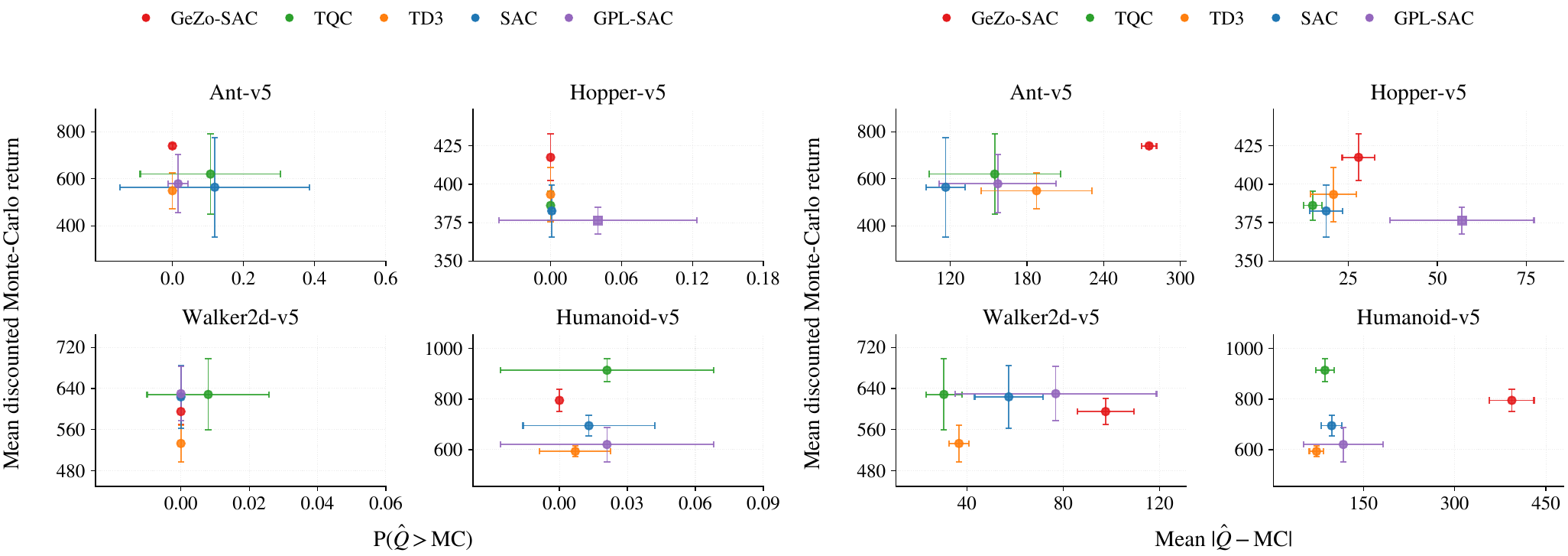}
\caption{\textbf{Value-estimation diagnostics and policy performance.}
Comparison of GeZo-SAC, SAC, TD3, TQC, and GPL-SAC across four
locomotion tasks.
\textbf{Left:} measured overestimation frequency
$P(\hat Q(s,a_\pi)>\mathrm{MC}(s))$.
\textbf{Right:} mean absolute critic--return discrepancy
$\mathbb{E}[|\hat Q(s,a_\pi)-\mathrm{MC}(s)|]$.
The vertical axis shows mean discounted, reward-only Monte-Carlo
return under deterministic evaluation.
Markers average policy-level metrics across training seeds;
horizontal and vertical error bars show $\pm$ one sample
standard deviation.}
    \label{fig:tradeoff}
\end{figure*}
\vspace{-1mm}
\section{Experiments}

\subsection{Experimental Setup}


\noindent\textbf{Baselines.}
We compare GeZo-SAC against six off-policy continuous-control methods:
\textit{SAC},
\textit{TD3},
\textit{TQC},
\textit{GPL-SAC},
\textit{TDC-$\lambda$}, and
\textit{SUNRISE}.
We use the same
environment and evaluation protocol across methods, while retaining
method-specific critic structures and aggregation mechanisms.

\noindent\textbf{Environments.}
We evaluate Ant-v5, Hopper-v5, Walker2d-v5, and Humanoid-v5,
four locomotion benchmarks from
Gymnasium~\cite{towers2024gymnasium}.

\noindent\textbf{Training.}
All experiments are conducted for $T = 3\times 10^6$ interaction steps per seed, with a maximum episode length of 1000. To ensure a fair and balanced evaluation, we use the same hyperparameters: discount factor $\gamma = 0.99$, Polyak averaging coefficient $\tau = 0.005$, Adam optimizer with learning rate $3\times 10^{-4}$, batch size 256 and buffer size $5\times 10^{5}$. 

\noindent\textbf{Evaluation Setup.}
For each method, we perform five independent training runs.
During training, policies are evaluated every 50 training
episodes using 10 deterministic evaluation episodes.
For each run, we select the checkpoint achieving the highest
mean return in these evaluations.
After training, the selected policy is evaluated over 300
deterministic episodes, using reset seeds different from
those used for checkpoint selection.

\noindent\textbf{Metrics.}
Beyond episodic return, we evaluate overestimation frequency,
critic--return discrepancy, and mechanical efficiency.

\noindent\emph{Value diagnostics.}
Overestimation frequency is measured as
$P(\hat Q(s,a_\pi)>\mathrm{MC}(s))$, where
$\mathrm{MC}(s)$ is the discounted Monte-Carlo return obtained
from the restored simulator state under the deterministic evaluation
policy. Lower values indicate less frequent optimistic estimates,
but do not by themselves establish greater estimation accuracy.
We therefore also report the mean absolute critic--return discrepancy
$
\mathbb{E}\!\left[
\left|\hat Q(s,a_\pi)-\mathrm{MC}(s)\right|
\right]
$
over the same subset of states.

\noindent\emph{Mechanical and control metrics.}
We evaluate locomotion efficiency using absolute actuator work per metre,
forward distance, action total variation (TV) per metre, and action effort per
metre. Let $\tau_t$ and $\dot q_t$ denote the generalized actuator force
and velocity vectors, and $\mathbf{a}_t$ the policy action at step $t$.
For an episode with forward displacement $\Delta x$, we compute
\begin{align}
\frac{E_{\mathrm{abs}}}{m}
&=
\frac{
\sum_{t=0}^{T-1}\sum_{d=1}^{n_v}
|\tau_t^{(d)}\dot q_t^{(d)}|\,\Delta t
}{
\max(|\Delta x|,\epsilon)
}, \\
\mathrm{TV}/m
&=
\frac{
\sum_{t=1}^{T-1}
\|\mathbf{a}_t-\mathbf{a}_{t-1}\|_1
}{
\max(|\Delta x|,\epsilon)
}, \\
\mathrm{Effort}/m
&=
\frac{
\sum_{t=0}^{T-1}
\|\mathbf{a}_t\|_2^2\,\Delta t
}{
\max(|\Delta x|,\epsilon)
}.
\end{align}
For the cross-environment summary, mechanical metrics are normalized within
each environment before averaging across tasks.

\subsection{Results}
\noindent\textbf{Performance and Training Dynamics.}
Table~\ref{tab:best300-7-methods} shows that GeZo-SAC achieves the
highest mean evaluation return on Ant-v5 and Hopper-v5, while remaining
competitive with the strongest baselines on Humanoid-v5 and Walker2d-v5.

Figure~\ref{fig:training_progress_panel} shows different learning patterns
across tasks. On Ant-v5, GPL-SAC improves fastest initially, but GeZo-SAC
overtakes it and remains among the strongest methods later in training.
On Hopper-v5, the leading methods show overlapping learning curves,
with GeZo-SAC achieving the highest mean return in the
selected-checkpoint evaluation.
On Humanoid-v5, TQC achieves the highest returns in later training,
with GeZo-SAC generally following ahead of the remaining methods.
On Walker2d-v5, GeZo-SAC reaches a comparable return range, although
its mean curve generally remains below the leading baselines. SUNRISE remains at low returns on Humanoid-v5, while TDC-$\lambda$ exhibits
a marked decline in return on Hopper-v5 and Walker2d-v5.



\noindent\textbf{Mechanical Cost and Control Quality.}
GeZo-SAC combines the highest average return with the lowest average
actuator work, action variation, and action effort per metre
(Table~\ref{tab:mechanical_averages}). It also ranks second in forward distance.

Figure~\ref{fig:oneplot_all_envs} illustrates trade-offs across environments. On Ant-v5, GeZo-SAC achieves the highest return with less work per metre than SAC, TD3, TQC, and TDC-$\lambda$. On Humanoid-v5, GeZo-SAC and TQC combine the highest returns with the lowest work per metre. On Walker2d-v5, GeZo-SAC and SUNRISE achieve similar returns at the lowest work per metre, whereas TQC and GPL-SAC obtain higher returns at greater mechanical cost. On Hopper-v5, GeZo-SAC reaches returns comparable to the leading methods, using less work per metre than TD3 and TDC-$\lambda$ but more than SAC and SUNRISE. The aggregate advantage therefore reflects different trade-offs across tasks, with GeZo-SAC offering particularly favorable combinations of return and mechanical cost on Ant-v5 and Humanoid-v5.

Figure~\ref{fig:ant_rollout} compares learned locomotion on Ant-v5. Synchronized poses show forward progress, while work–displacement curves track mechanical cost. Over the illustrated one-second interval, GeZo-SAC travels farther than TQC and GPL-SAC while expending less absolute actuator work, achieving the lowest work per metre of the three displayed rollouts.

\subsection{Value Estimation and Policy Performance}
Figure~\ref{fig:tradeoff} compares GeZo-SAC with SAC, TD3, TQC,
and GPL-SAC using the mean online critic prediction for each method,
without the pessimistic adjustments used during training.
Metrics are computed over sampled states from the deterministic
evaluation policy, and error bars show variability across training seeds.

GeZo-SAC exhibits near-zero measured overestimation frequency across
all four environments. On Ant-v5, it combines the highest mean
Monte-Carlo return with lower measured overestimation frequency than
SAC and TQC. On Hopper-v5, GeZo-SAC, SAC, TD3, and TQC show near-zero
measured overestimation frequency, while GPL-SAC has a higher mean
frequency. GeZo-SAC attains the highest mean Monte-Carlo return. On Humanoid-v5, TQC
achieves a higher return despite more frequent overestimation.

The absolute discrepancies reveal a different ordering. GeZo-SAC
has the largest mean critic--return discrepancy on Ant-v5,
Humanoid-v5, and Walker2d-v5, despite achieving the highest return
on Ant-v5 and the second-highest on Humanoid-v5. Absolute agreement
with Monte-Carlo returns therefore does not directly predict policy
quality. An action-independent offset, for example, changes absolute
value errors without changing action preferences. On Walker2d-v5,
however, several baselines combine smaller discrepancies with higher
returns. 

\subsection{Ablation: Number of Zonotope Generators}
We vary the number of generators
$K\in\{1,2,4,6,8,10,15,20\}$ across all four environments, keeping
the remaining hyperparameters fixed and using five independent
training seeds per configuration.

Figure~\ref{fig:zono_comparison} shows that the effect of $K$ is
task-dependent. Ant-v5 and Hopper-v5 attain their highest mean returns
with larger generator sets, at $K=15$ and $K=20$, respectively.
Larger configurations also generally show less variability across
seeds on these tasks. In contrast, Humanoid-v5 and Walker2d-v5
achieve their highest mean returns at $K=1$ and $K=2$, with no
consistent benefit from additional generators.
All main comparisons use a shared setting of $K=15$ across the
four environments.

\begin{figure*}[t]
    \centering
    \includegraphics[width=\textwidth]
        {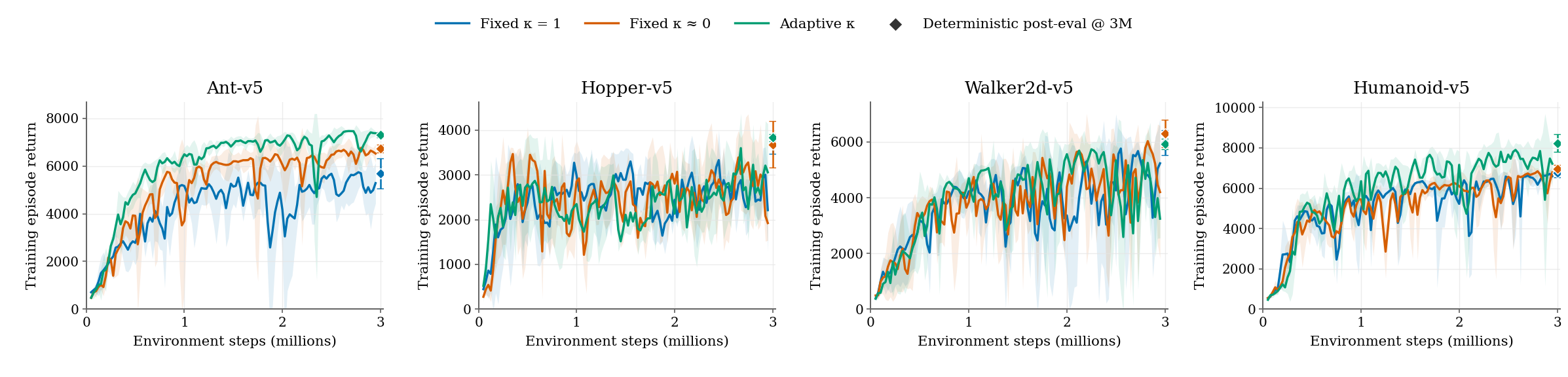}
\caption{\textbf{Effect of fixed versus adaptive pessimism.}
Evaluation returns for GeZo-SAC with fixed $\kappa=1$, fixed
$\kappa\approx 0$, and the adaptive $\kappa$ controller across the four
locomotion tasks. Curves show the mean over three training runs, with
shading indicating $\pm$ one sample standard deviation. Diamonds mark
the final deterministic evaluation over 300 episodes per policy, with
error bars showing the standard deviation across runs.}
    \label{fig:kappa_returns}
\end{figure*}

\begin{figure}[t]
    \centering
    \includegraphics[
    width=.65\linewidth,
]{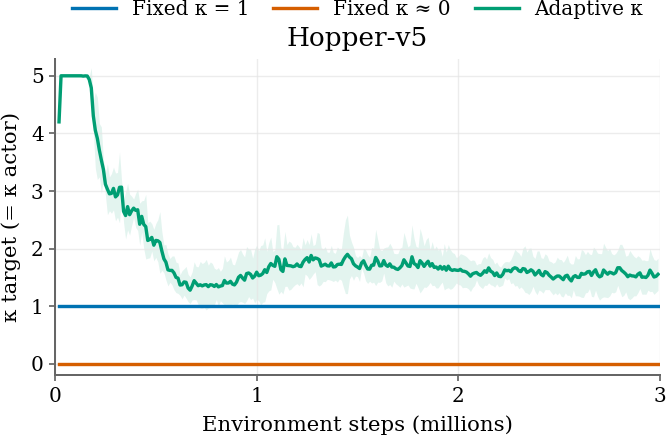}
\caption{\textbf{Evolution of the adaptive pessimism coefficient on Hopper-v5.}
The adaptive controller changes $\kappa$ throughout training, while the two
ablated variants keep it fixed. The curve shows the mean across three training
seeds, with shading indicating $\pm$ one standard deviation.}
    \label{fig:kappa_progress}
    \vspace{-0.35cm}
\end{figure}


\subsection{Ablation: Fixed versus Adaptive Pessimism}
We compare fixed $\kappa=1$, near-zero $\kappa=10^{-6}$,
and adaptive $\kappa$, using $K=15$. Each setting is trained
for $3\times10^6$ environment steps with three independent seeds
per environment
(Figs.~\ref{fig:kappa_returns}--\ref{fig:kappa_progress}).
In the displayed runs, adaptive $\kappa$ achieves higher returns
on Ant-v5 and Humanoid-v5. Hopper-v5 shows overlapping training
curves but the highest final evaluation return with adaptation,
whereas near-zero $\kappa$ performs best on Walker2d-v5.
Figure~\ref{fig:kappa_progress} also illustrates the evolution of the adaptive
pessimism coefficient on Hopper-v5. Starting from its initial value, the
controller decreases $\kappa$ substantially during early training and
subsequently stabilizes at an intermediate level, in contrast to the fixed
coefficients used in the ablations.

\begin{table}[t]
\centering
\footnotesize
\setlength{\tabcolsep}{2.2pt}
\begin{tabular}{lrrrr}
\toprule
Method & Act. (M) & Crit. (M) & Train (ms) & Infer. (ms) \\
\midrule
SAC & 0.101 & 0.199 & 9.83 & 0.181 \\
TD3 & 0.099 & 0.199 & 5.25 & 0.161 \\
TQC & 0.101 & 0.527 & 16.90 & 0.181 \\
GPL-SAC & 0.101 & 1.656 & 41.62 & 0.662 \\
TDC-$\lambda$ & 0.099 & 0.231 & 6.65 & 0.153 \\
SUNRISE & 0.506 & 0.993 & 55.63 & 0.900 \\
GeZo-SAC (ours) & 0.101 & 1.133 & 15.71 & 0.181 \\
\bottomrule
\end{tabular}
\caption{\textbf{Computational overhead.}
Parameter counts (in millions, M) include the actor and all online critics and are averaged over the four environments. Timings calculated on an NVIDIA A100-SXM-80GB}
\label{tab:compute-8}
\vspace{-0.4cm}
\end{table}


\subsection{Computational Cost}

Table~\ref{tab:compute-8} compares model size and runtime across methods.
GeZo-SAC increases critic capacity relative to SAC, but its training cost
remains comparable to TQC and substantially below larger ensemble-based
methods such as GPL-SAC and SUNRISE. Importantly, the additional zonotope
representation is confined to the critics. The deployed actor therefore has
the same architecture and inference latency as SAC, so the extra computational
cost is incurred only during training.



\section{Conclusions}
We presented GeZo-SAC, which augments SAC's twin critics with auxiliary
zonotope generators whose directional supports provide heuristic
geometric bands and a disagreement signal for adapting pessimism and
critic aggregation. Across four MuJoCo-v5 tasks, GeZo-SAC achieves the
highest mean return on Ant-v5 and Hopper-v5, outperforms SAC and TD3 on
every task, and attains the lowest average work and action effort per
metre, with near-zero measured overestimation frequency. In the future we also plan to extend and evaluate GeZo-SAC to robotic manipulation tasks.

\bibliographystyle{IEEEtran}
\bibliography{refs}

\onecolumn          
\appendix
\section{Illustrated Overview of Geometric Critic Aggregation}
\label{app:geometric_overview}

\begin{figure}[H]
    \centering
    \includegraphics[width=\textwidth]{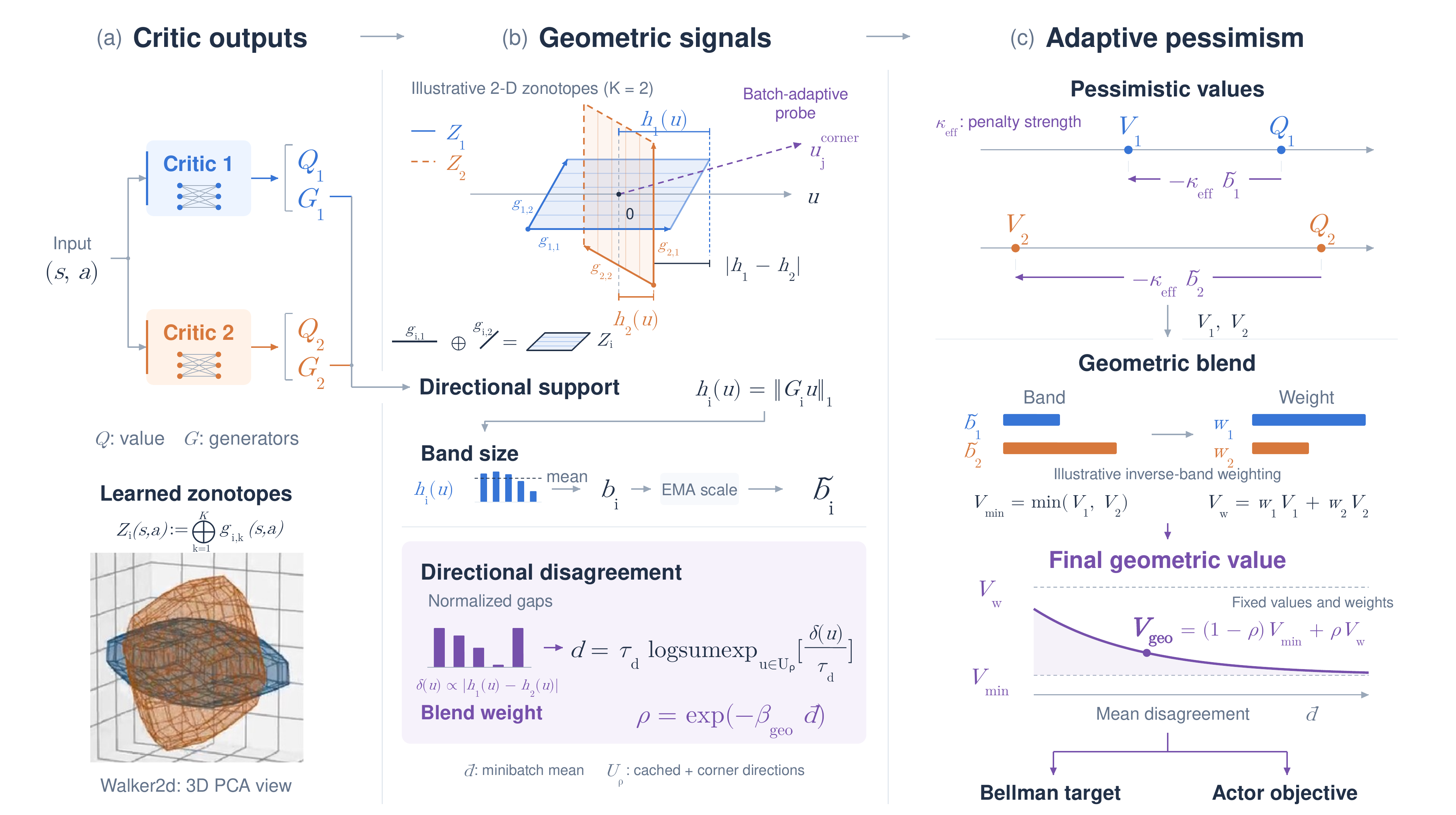}
    \caption{
    Geometric critic aggregation in GeZo-SAC.
    (a) Each critic predicts a scalar value and a set of
    state--action-conditioned zonotope generators.
    (b) Directional supports provide geometric bands and
    cross-critic disagreement.
    (c) The bands determine pessimistic value offsets and
    inverse-band weights, while disagreement controls the
    blend between the weighted value and the minimum.
    The Walker2d inset shows a 3D Principal Component Analysis (PCA) visualization of the learned zonotopes;
    the planar geometry and miniature plots are illustrative.
    }
    \label{fig:geometric_overview}
\end{figure}

Figure~\ref{fig:geometric_overview} illustrates the main
geometric aggregation mechanism of GeZo-SAC. It follows
the information used to construct the final geometric value,
rather than depicting the complete training loop.
For readability, dependence on the state--action pair
$(s,a)$ is omitted below.

\noindent\textbf{(a) Critic outputs.}
Each critic predicts a scalar value $Q_i$ and generators
$G_i$ defining an auxiliary zonotope. The directional support $h_i(u)=\|G_i u\|_1$
measures the maximum extent of the zonotope
when projected onto the unit direction $u$.
The Walker2d PCA inset shows learned geometry; the
planar drawing is a separate illustrative schematic.

\noindent\textbf{(b) Geometric signals.}
Averaging supports over cached directions gives the
band $b_i$, which is EMA-normalized to $\widetilde b_i$.
Normalized differences between the two zonotopes'
directional supports measure geometric disagreement.
These differences are evaluated along stored directions
and additional directions computed from the minibatch. Temperature-LSE emphasizes
larger gaps, and its minibatch mean $\overline d$
determines $\rho=\exp(-\beta_{\mathrm{geo}}\overline d)$.
Additional probe directions are obtained from
minibatch-averaged differences between corresponding
generators of the two critics.

\noindent\textbf{(c) Adaptive pessimism.}
The bands produce pessimistic values
$V_i=Q_i-\kappa_{\mathrm{eff}}\widetilde b_i$,
where $\kappa_{\mathrm{eff}}$ controls the strength
of the geometric penalty.
Normalized inverse-band weights give greater weight
to the critic with the smaller band. The final value is
\[
V_{\mathrm{geo}}
=(1-\rho)\min(V_1,V_2)+\rho(w_1V_1+w_2V_2).
\]
For fixed values and weights, greater disagreement
moves this blend toward the minimum.
Aggregation is computed separately for the Bellman
target and actor objective.

\end{document}